# Segmentation and Classification of Skin Lesions for Disease Diagnosis

Sumithra R[a], Mahamad Suhil[b], D.S.Guru[c]

*[a,b,c] Department of Studies in Computer Science, University of Mysore, Mysore, India.*

**Abstract**

In this paper, a novel approach for automatic segmentation and classification of skin lesions is proposed. Initially, skin images are filtered to remove unwanted hairs and noise and then the segmentation process is carried out to extract lesion areas. For segmentation, a region growing method is applied by automatic initialization of seed points. The segmentation performance is measured with different well known measures and the results are appreciable. Subsequently, the extracted lesion areas are represented by color and texture features. SVM and k-NN classifiers are used along with their fusion for the classification using the extracted features. The performance of the system is tested on our own dataset of 726 samples from 141 images consisting of 5 different classes of diseases. The results are very promising with 46.71% and 34% of F-measure using SVM and k-NN classifier respectively and with 61% of F-measure for fusion of SVM and k-NN.

## 1. Introduction

Skin is the largest organ in a human body, which helps to cover the muscles, bones and all parts of the body. Functions of skin in human body have greater importance because; a small change in its functioning might affect other parts of the body. Skin is exposed to outer environment thus the disease and infection occurs more to skin. So, we have to give a greater attention to skin disease. The spot on skin which is infected is called a lesion area. Skin lesions are the first clinical signs of disease such as chickenpox, Melanoma, etc. Nowadays, medical field depends more on computer-aided diagnosis. Early detection of skin disease is more complex to the inexperienced dermatologist. By incorporating digital image processing for skin cancer detection, it is possible to do the diagnosis without any physical contact with skin. For these reasons, developing Computer-Aided Diagnosis System (CADs) has become a major area of research in the medical field. Machine learning plays an essential role in the medical field for the automation of many processes. It has been demonstrated that dermoscopy may actually lower the diagnostic accuracy in the hands of inexperienced dermatologists.

Various skin lesion classification systems have been proposed in the literature. Abbas et al., (2013) explain that a pattern classification of dermoscopy images for a perceptually uniform model. Celebi et al., (2013) presents a methodological approach to the classification of pigmented skin lesions in dermoscopy images. Yuan et al., (2008) presented an evolution strategy (ES) based segmentation algorithm to identify the lesion area within an ellipse. Rajab et al., (2003) proposed two approaches to the skin segmentation problem. First is region-based segmentation method where an optimal threshold is determined iteratively by using iso-data algorithm. The second method proposed is based on neural network edge detection method. Wang et al., (2010) proposed watershed-based algorithm for automatic segmentation in dermoscopy images. Wighton et al., (2010) have presented a method for automatically segmenting skin lesions by initializing the random walker algorithm. Li et al., (2010) use 3D depth information to RGB colour images to improve segmentation of pigmented and non-pigmented skin lesion. Komati et al., (2010) presented an improved version for the JSEG color image segmentation algorithm, combining the classical JSEG algorithm with local fractal operator. Elgamal and Uni (2013) proposed that early detection of skin cancer has the potential to reduce mortality and morbidity. Shena et al., (2012), present an automated method for melanoma diagnosis applied on a set of dermoscopy images using texture analysis. Aswin et al., (2013) propose a computer aided classification system using ANN classifier. Mahmoud et al., (2013) presents an automatic skin cancer classification system using neural network classifier with wavelet and curvelet based features.

From the literature, we can observe that, majority of the works consider only 3 types of diseases namely melanoma, squamous cell and seborrheic keratosis. But there are some other diseases which possess equal importance in the medical field. So we consider 2 such additional diseases namely bullae and shingles along with the previously mentioned diseases. Another important aspect is the features used to describe the lesion regions. We consider color and texture features for this purpose. One more point that needs to be observed is the learning algorithms. Although, many classifiers are used, fusion of decisions from multiple classifiers is recently gaining importance due to the fact that improvement in classification can be achieved. So, in our work we have fused the results of SVM and k-NN classifiers.

The rest of the paper is organized as follows. Section 2 presents the dataset description. In section 3 the proposed model is presented. Experimentation and Results are explained in Section 4. Finally, section 5 conclusions.

## 2. Dataset Description

Creation of a suitable dataset is crucial for any work as the dataset is a primary thing required for designing and testing the system. Due to the unavailable benchmarking dataset and permission issues we have created our own dataset by downloading images from internet resources. These were true-color images with varying resolution. Since, we had no control over the image acquisition and camera calibration, and some images within a class have different contrast and illumination. As, images are collected through internet, they exhibit a large intraclass variations with less interclass variations. This sensitivity was necessary in order to ensure accurate pre processing technique such as filtering. In this work, we select five different classes of skin diseases: Melanoma, bullae, seborrheic keratosis, shingles and squamous cell. Figure 2.1 shows some of the samples from these five different classes of skin diseases. A total of 141 images are collected out of which 31 are melanoma, 26 are bullae, 33 are seborrheic keratosis, 20 are shingles, and 31 are squamous cell. We can observe from the figure 1, that, images within a class are with high variation and also it is quite difficult to distinguish between images from different classes due to the similarity between the classes. One skin image may have more than one lesion area. We have considered every lesion area as a separate sample after segmentation, so that the collection size is increased to 726 samples.

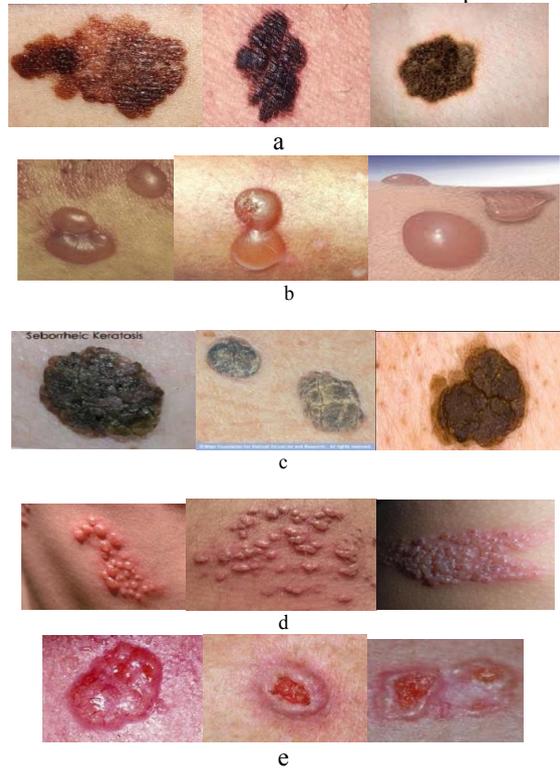

Fig. 1. (a) Melanoma; (b) Bullae; (c) Seborrheic keratosis; (d) Shingles; (e) Squamous cell

## 3. Proposed Model

The proposed methodology is to design and develop a computer vision based system for segmentation and classification of skin lesions along with extraction of dicriminating set of features from skin lesions for efficient classification. The overview of the proposed method is shown in the figure 2 below.

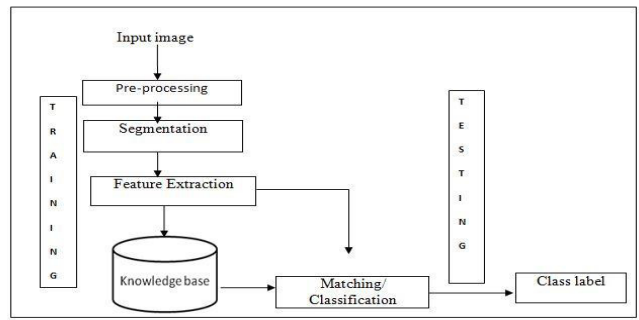

Fig. 2. Overview of proposed method for skin lesions classification.

### 3.1. Pre-processing and Segmentation

The first step in the computerized analysis of skin lesion images is the pre-processing of an image. The pre-processing techniques will be different for different application based on the desired dataset of an image. The main aim of pre-processing techniques is image enhancement and image restoration [1, 5]. Due to noise in image we conduct filtering technique called Gaussian smoothing [10]. In this work, we employed 3×3 gaussian filter to smoothen the image because the gaussian smoothing in 2D convolution operation is used to 'blur' images nd remove hair and noise (figure 3 (a) and (b)).

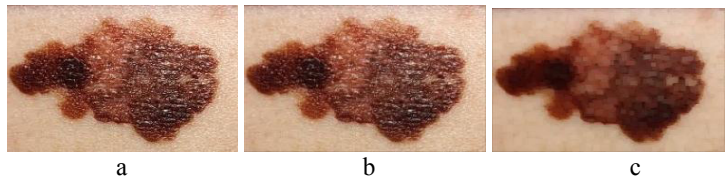

Fig. 3. (a) Input image; (b) Filtered Image after Gaussian filtering; (c) Filtered image after Morphological operations

Further, to enhance the filtered image, appling morphological erosion and dilation. Figure 3 (c) shows the image after morphological operations.

Segmentation subdivides an image into its constituent regions or objects. That is, segmentation should stop when the objects of interest in an application have been isolated. There are many method for segmentation: watershed based segmentation, split and merge, region growing, threshold based segmenation etc,. In this work, we have used region growing method [2, 3, 6], also called pixel-based image segmentation then decided initial seed points emperically. Some of the segmentation results are shown in figure 3 by observing the segmentation result is good with less amount of over segmentation and under segmentation of lesion area.

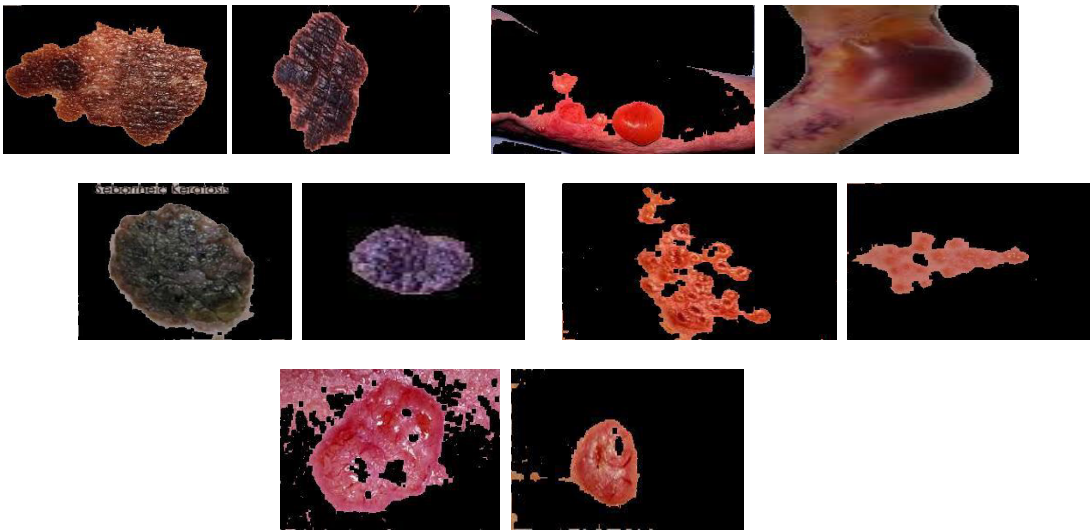

Fig. 4. Sample results of segmentation

### 3.2 Feature Extraction

In this section the features which have been used to characterize the skin lesion images are described. In this work, we use color, texture, and color histogram features to represent lesion areas. The reason for choosing these types of features is because of the fact that color and texture are the only properties dominating in the lesion region

*3.2.1. Color Features*

In order to quantify the color present in a lesion, we extract four statistics (mean, standard deviation, variation and skewness) from segmented lesion regions over individual channels of six different color spaces : *RGB, HSV, YCbCr,*

*NTSc, CIE L\*u\*v and CIE L\*a\*b* [3, 7, 8]. These four color feature are also termed as color moments.

Let $P_j$ be the j[th] pixel of a color channel P of an image i with N pixels in a color space. Then, the four color moments are defined as follows

Moment 1- Mean is the average color value in the channel which is given by,

$$\mu = \frac{1}{N} \sum_{j=1}^{N} p \tag{1}$$

Moment 2- Standard deviation is the square root of the variance of the distribution which is given by,

$$\sigma = \sqrt{\frac{1}{N} \sum_{j=1}^{N} (p_j - \mu)^2} \tag{2}$$

Moment 3- Skewness is the measure of the degree of asymmetry in the distribution which is computed by,

$$S = \sqrt[3]{\frac{1}{N} (\sum_{j=1}^{N} (p_j - \mu)^3)} \tag{3}$$

Moment 4- Variance is the variation of the color distribution. Given by.

$$V = \frac{1}{N} \sum_{j=1}^{N} (p_j - \mu)^2 \tag{4}$$

The above described four features are computed over every channel which results in 72 color features obtained by the following combination: (6 color spaces)×(3 channels in each color space)×(4 features).

### 3.2.2. Texture Features

In order to quantify the texture present in a lesion, a set of statistical texture descriptors based on the gray level co-occurance matrix (GLCM) were employed [4, 8]. GLCM-based texture descriptions are well-known and widely used methods for texture computation in the literature[11]. Co-occurrence texture features are extracted from the image in two steps. First, the pair wise spatial co-occurrence of pixels separated by an exact angle and a distance are developed using a gray level co-occurrence matrix (GLCM). Second, the GLCM is used to calculate a set of scalar quantity that distinguish different aspects of the original texture. The GLCM is a M×M square matrix, where M is the number of different gray levels in an image. An element $X(i,j,d,\theta)$ of a GLCM of an image represents the relative frequency, where i is the gray level of the pixel X at location(x,y), and j is the gray level of a pixel located at a distance d from X in the orientation θ. It calculates how a pixel with gray-level i occurs horizontally adjacent to a pixel with the value j. Haralick et al, (1973) proposed a set of scalar quantities for summarizing the information contained in a GLCM. Originally proposed a total of 14 features namely, Angular second moment, contrast, correlation, sum of variance, inverse difference moment, sum average, sum variance, sum entropy, entropy, difference variance, difference entropy, information measures of correlation, and maximal correlation coefficient. In order to obtain texture features [9], the normalized GLCM was computed for each of the four orientations ($\{0^0, 45^0, 90^0, 135^0\}$). All 14 features are computed for each orientation individually, and a feature vector of 14 dimensions is created by taking the average of the features of individual orientation.

### 3.2.3. RGB Histogram Features

An image histogram is a graphical representation of the tonal distribution in a digital image. It plots the number of pixels for each tonal value. The horizontal axis of the graph represents the tonal variations, while the vertical axis represents the number of pixels in that particular tone. The color histogram is a method for describing the color content of an image; it counts the number of occurrences of each color in an image. In order to obtain RGB histogram features the number of bins is set to 16 for individual sub bands. The total number of bins due to all 3 components is equal to $16^3$ which is 4096. Therefore every sample is represented by a RGB histogram of size 4096.
Overall, the number of features extracted from each lesion area is 4182 (72 color, 14 texture, 4096 RGB histrogram features).

## 4. Experimentation and Results

*4.1 Segmentation Performance Analysis*

To measure the performance of segmentation, we made a setup by marking the lesion area called ground truth by the guidance of an expert dermatologist then compared the results of segmented image with the ground truth using the well-known segmentation performance measures proposed[12]. They are as follows,

Measure of Overlap (MOL): It measures the overlapping area between segmented and ground truth area. The probability of segmentation performance is superior, when MOL is high.

Measure of Under Segmentation (MUS): It measures the percentage of under segmentation of segmented ground truth area. The segmentation performance is superior, if MUS is low.

Measure of Over Segmentation (MOS): It measures the percentage of over segmentation of segmented ground truth area. Lower the MOS, superior is the segmentation performance.

Dice Similarity Measure (DSM): It measures similarity between segmented areas with ground truth area. If DSM is high, segmentation is superior.

Along with the above, we also make use of the segmentation measures proposed[13] called as Error Rate (ER) which measures the error in segmentation. If segmentation is said to be good if the ER is less.

Figure 5 gives the class-wise segmentation results using the performance measures described above. We can observe from that the segmentation performance is quite good for the melanoma and seborrheic keratosis. If reasonably bad segmentation resulted in bullae and shingles because of less variation between skin and lesion areas.

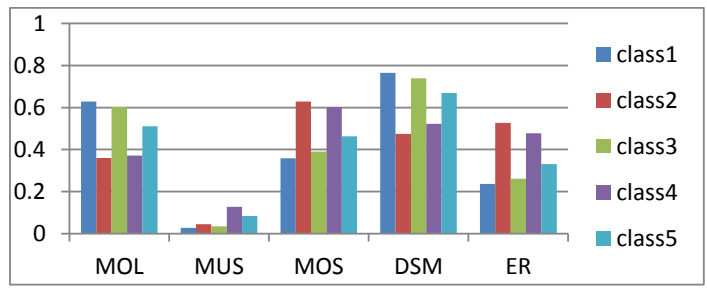

Fig. 5. Segmentation results of individual class

*4.2 Classification performance analysis*

Once features are extracted, a classifier can be trained to classify a test sample as a member of one of the known classes. In literature [14], we can find a number of classifiers including both parametric and non-parametric type. For our experimentation we use two well-known classifiers namely, linear Support Vector Machine (SVM) and k- Nearest Neighbour (k-NN) classifiers. We also use fusion of decisions of both SVM and k-NN by using OR rule.

To measure the performance of classification, we conducted five different sets of experiments. In the first set of experiments, we used 30% of the samples of each class of the dataset to create class representative vectors (training) and the remaining 70% of the samples for testing purpose. Remaining set of experiments are with the number of training and testing samples in the ratio 40:60, 50:50, 60:40, and 70:30 respectively. In each set, experiments are repeated 20 times by choosing the training samples randomly. For k-NN classifier we decided the k value empirically. During experimentation, we conduct experiment of 20 trials for every set of training and testing samples randomly.

The performance of the classification is evaluated in terms of classification accuracy, precision, recall and F-measure from the confusion matrix of classification. The measures are computed by using the equations described below with the following conventions.
TP (True Positive) =Positive samples classified as positive. TN (True Negative) = Negative samples classified as negative. FP (False Positive) = Negative samples classified as positive. FN (False Negative) = Positive samples classified as negative.

**Precision**: It is the ratio of number of positive samples correctly classified to the total number of samples in a class.

$$\text{Precision} = \frac{TP}{(TP+FP)} \quad (5)$$

**Recall**: It is the ratio of number of positive samples correctly classified to the total number of samples classified as positive.

$$\text{Recall} = \frac{TP}{(TP+FN)} \quad (6)$$

**F-measure**: It is the harmonic mean of precision and recall given by the equation below.

$$\text{F-measure} = \frac{(2 * \text{Precision} * \text{Recall})}{(\text{Precision} + \text{Recall})} \quad (7)$$

**Accuracy**: It is the total number of samples correctly classified to the total number of samples classified as given by the equation below.

$$\text{Accuracy} = \frac{(TP + TN)}{(TP+TN+FP+FN)} \quad (8)$$

The performance of the classifiers is with 20 different trials are presented in the figures below. Minimum, average and maximum accuracy of SVM, k-NN and SVM+KNN fusion classifiers under varying percentage of training samples for each individual class with 70 percent of training samples is shown in the Figure 4.2.

It is observed from Table 4 that, the classification performance for both SVM and k-NN is high for 70 percent of training samples. However, SVM classifier has achieved a better performance with 46.71 percent of F-measure where as k-NN classifier has given only 34.1 percent of F-measure for 70% of training samples. Maximum performance is obtained by the fusion of SVM with k-NN with 61.03 of F-measure for the same 70 percent of training samples.

To compare the performance of the classifiers on individual classes we have given the class wise F-measure using SVM, k-NN and SVM+ k-NN fusion with 70 percent of training samples. It can be observed from the Table 4.2 that, all the classifiers are showing very good performance for the class 3 i.e., Seborrheic keratosis. This might be because of less intra class variations and the effective segmentation. The performance of the remaining classes are poor which might be because of various reasons such as more intra class variations, less interclass variations, poor segmentation, unsuitability of the extracted features etc., which we will try to sort out in our future work.

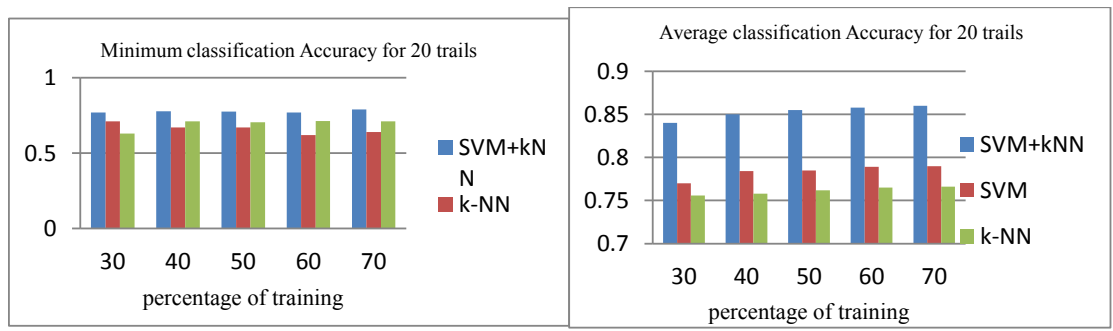

(a)  (b)

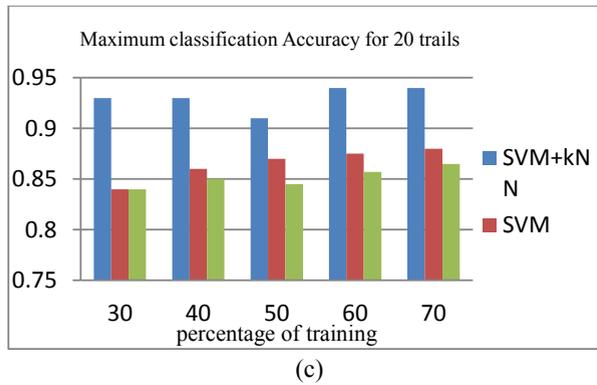

(c)

Fig. 6. Accuracy for different classifier combination (a) Minimum; (b)Average; (c) Maximum;

Table 1. Average F-measure with various classifier for varies percentage of training.

| Percentage of training samples | SVM | k-NN | SVM +k-NN |
| --- | --- | --- | --- |
| 30 | 0.47 | 0.29 | 0.57 |
| 40 | 0.43 | 0.31 | 0.58 |
| 50 | 0.43 | 0.32 | 0.59 |
| 60 | 0.45 | 0.33 | 0.6 |
| 70 | **0.46** | **0.34** | **0.61** |

Table 2. Average F-measure of different classes with various classifiers.

| Classifiers | Melanoma | Bullae | Seborrheic keratosis | Shingles | Squamous cell |
| --- | --- | --- | --- | --- | --- |
| SVM | 0.397 | 0.44 | 0.646 | 0.435 | 0.427 |
| k-NN | 0.323 | 0.238 | 0.618 | 0.242 | 0.296 |
| SVM + k-NN | 0.53 | 0.531 | 0.861 | 0.565 | 0.581 |

In order to examine the superiority of the proposed method, a qualitative comparative study with other contemporary methods is presented in Table 5.1. According to comparison made, we can see that our method is superior to other state of the art techniques in many respects.

## 5. CONCLUSION

In this work, a model for the segmentation and classification of skin lesionsis proposed. Initially lesion areas are segmented using region growing method. Later, Color and texture features are extracted to represent segmentedlesion areas. Then the classification is performed with SVM, KNN and fusion of SVM and KNN Classifiers. Experiments are conducted on our own dataset of 726 lesion samples from 5 different classes of skin diseases. Among the 5 types of diseases considered, two are nowhere dealt for classification even though they are of equal clinical importance as others. The computed performance measures of the classification show that the SVM-KNN fusion based classifier is outperforming individual classifiers. A comparative study is also performed with other contemporary works to check the superiority of the proposed method. Hence, it is revealedfrom the results of the classification that the proposed method can be used as a supplementary tool for the experts to diagonise skin diseases.

Based on the performance of the proposed model we have observed the performace of the system has descreased quite considerably for some classes and because of which the overall performance is also affected. This is due to a collection of dataset from internet resources. We will think of experimenting with datasets of collection from hospitals. We are also thinking to use some of the well known feature selection methods to select good features, to improve performance of system. Further,the improvement in classification might also be achieved through the use of different classifiers such as ANN classifier, PNN classifier etc., along with their combination. So, in our future work we also think of working with fusion at different levels such as feature level, decision level etc.,

Table 3. Qualitative comparision of the proposed model with other well known models for segmentation and classification of skin lesions.

| Title | Classes | Segmentation method | Features | Classifiers |
|---|---|---|---|---|
| A methodological approach to the classification of dermoscopy images | Benign and Melanoma 2 classes | JSEG algorithm for segmentation | 1. shape features: area, asymmetry, compactness and diameter. 2. Color features: color space of color moment- mean, standard deviation, centroid distance. 3. Texture feature: GLCM. | 1. Support vector Machines- decision trees and neural networks. 2 k-Nearest Neighbour. |
| Computer Aided Diagnosis of Mealanocytic lesions | Melanomas and Mealanocytic. 2 classes | Segmentation by morphological operation. | 1.Color 2.Texture | A termed statistical learning support vector machines (SVM) |
| Pattern classification of dermoscopy images: a perceptually uniform model | Benign Melanocytic lesion and Melanoma. 2 classes | Region of interest are extracted. | color symmetry and multiscale texture features are extracted using Color-texture properties. | Adaptive Boosting Multi channel (AdaBoostMC) classification. |
| Automated Melanoma Recognisation. | Benign, Dysplastic lesion and Melanoma. 3 classes | - | size: area, permeter, ploar measures and bounding rectangle. Shape: roundness, compactness. Color : color spaces of color moment . Gradient features | k-NN classifier |
| Our proposed model on Segmentation and Classification of skin lesions for disease diagnosis. | Melanoma, Bullae, Seborrheic keratosis, Shingles and Squamous cell 5 classes | Region growing segmentation is developed | Color features: color space with color moment and RGB histogram. Texture features: GLCM with Haralick features. | SVM, k-NN and combined SVM and k-NN classifiers are developed. |